\documentclass[10pt, a4paper]{article}
\usepackage{lrec}
\usepackage{multibib}
\newcites{languageresource}{Language Resources}
\usepackage{graphicx}
\usepackage{tabularx}
\usepackage{soul}
\usepackage{titlesec}
\titleformat{\section}{\normalfont\large\bf\center}{\thesection.}{1em}{}
\titleformat{\subsection}{\normalfont\SmallTitleFont\bf\raggedright}{\thesubsection.}{1em}{}
\titleformat{\subsubsection}{\normalfont\normalsize\bf\raggedright}{\thesubsubsection.}{1em}{}
\renewcommand\thesection{\arabic{section}}
\renewcommand\thesubsection{\thesection.\arabic{subsection}}
\renewcommand\thesubsubsection{\thesubsection.\arabic{subsubsection}}

\usepackage{epstopdf}
\usepackage[utf8]{inputenc}

\usepackage{hyperref}
\usepackage{xstring}

\usepackage{color}

\usepackage{xcolor}
\newcolumntype{b}{>{\hsize=.55\hsize}X}
\newcolumntype{a}{>{\hsize=.5\hsize}X}
\newcolumntype{s}{>{\hsize=.15\hsize}X}
\newcolumntype{y}{>{\hsize=.2\hsize}X}
\newcolumntype{t}{>{\hsize=.18\hsize}X}
\newcolumntype{z}{>{\hsize=.10\hsize}X}
\newcolumntype{e}{>{\hsize=.4\hsize}X}

\title{Recognition of Implicit Geographic Movement in Text}

\name{Scott Pezanowski, Prasenjit Mitra}

\address{Information Sciences and Technology \\
The Pennsylvania State University \\
University Park, PA, USA 16802 \\
\{scottpez, pmitra\}@psu.edu\\}

\abstract{
Analyzing the geographic movement of humans, animals, and other phenomena is a growing field of research. This research has benefited urban planning, logistics, animal migration understanding, and much more. Typically, the movement is captured as precise geographic coordinates and time stamps with Global Positioning Systems (GPS). Although some research uses computational techniques to take advantage of implicit movement in descriptions of route directions, hiking paths, and historical exploration routes, innovation would accelerate with a large and diverse corpus. We created a corpus of sentences labeled as describing geographic movement or not and including the type of entity moving. Creating this corpus proved difficult without any comparable corpora to start with, high human labeling costs, and since movement can at times be interpreted differently. To overcome these challenges, we developed an iterative process employing hand labeling, crowd voting for confirmation, and machine learning to predict more labels. By merging advances in word embeddings with traditional machine learning models and model ensembling, prediction accuracy is at an acceptable level to produce a large silver-standard corpus despite the small gold-standard corpus training set. Our corpus will likely benefit computational processing of geography in text and spatial cognition, in addition to detection of movement. \\ \newline \Keywords{geographic movement, spatial cognition, text classification, crowdsourcing, corpus, machine learning, embeddings} }

\begin{document}

\maketitleabstract

\section{Introduction}

The study of movement of humans, animals, and other entities throughout geographic space has a large and growing body of research~\cite{Dodge2012,Dodge2016,Dodge2016a,Dodge2016b,Gonzalez2008,Huang2017,Soares2017}. Datasets exist that allow scientists to produce valuable knowledge about these movements to improve urban planning, better understand animal migrations, and detect unusual movements of vessels. However, text sources that contain descriptions of movement also exist such as emails, social media posts, web documents, written historical documents, and other sources, and these sources are currently underutilized. In fact, IBM recently estimated that 80\% of the World's data comes in the form of unstructured data \cite{ibm2016unstructured} and perhaps the biggest form of unstructured data is text. Although data produced by GPS is becoming more common, undergoing the process of using GPS to collect large-scale data is still expensive for many use-cases and therefore commonly does not exist for many entity types. For text sources that are identified as containing descriptions of movement, they can be converted into a traditional geographic data format using Geographic Information Retrieval (GIR) which automatically detects place mentions in text and resolves those place mentions to the intended place in the World, thereby allowing for further analysis.

In order to make better use of this underutilized information source, we created a corpus of statements that describe geographic movement at both a small gold-standard level verified by humans \cite{pez2020movementlr} and at a large high quality silver-standard level. We foresee our corpus benefiting three different overall fields of research. First, most directly it can help improve the detection of statements of movement thereby making better use of this important but underused source of information to study geographic movement. Machine learning models can be trained on our corpus and used to predict and find other such statements. For a computer to be able to differentiate between the statement describing movement \textit{Hawks migrate from Nova Scotia, through Pennsylvania, to Georgia} and the somewhat similar sentence that does not describe movement \textit{Some salmon live in the Pacific Ocean while other salmon live in freshwater inland lakes}, a correctly labeled corpus is required.

Second, we anticipate that GIR itself will benefit from our corpus. For a computer to understand and resolve a place mention such as in the first statement above which is about the state of Georgia in the United States and not the country of Georgia, GIR techniques require many samples that have valuable contextual information and other co-occurring place mentions \cite{ju2016things}. As \newcite{maceachren2014place} hypothesized in his position statement, GIR would be aided by a greater focus on the context around place mentions in text in order to determine if 1) the word(s) are actually intended as a place mention or if it is not a place mention but another entity like the name of a person (Virginia) or part of a disease name (West Nile Virus) and 2) the author's intended place like London, Canada instead of London, England. Our corpus will provide a labeled dataset with many place mentions, co-occurring place mentions, and context in the statements about the place mentions and their relationships with other place mentions. This annotated data can be used to train many different geospatially-aware applications.

Third, our corpus can provide spatial cognition researchers more resources to examine how people think, write, and interpret geographic movement. An example of future work with our corpus in spatial cognition is to have humans classify the statements into different types of descriptions of movement based on their own cognitive understanding. These types could then improve our understanding of how people interpret movement; with this knowledge feeding back into GIR as hints to use the context and co-occurring place mentions to resolve place mentions to the correct location. Also, other applications can be improved where this knowledge is important such as producing automated descriptions of route directions in online routing services in the most easily understood way.

\begin{figure}[!htb]
\begin{center}
\includegraphics[width=\columnwidth]{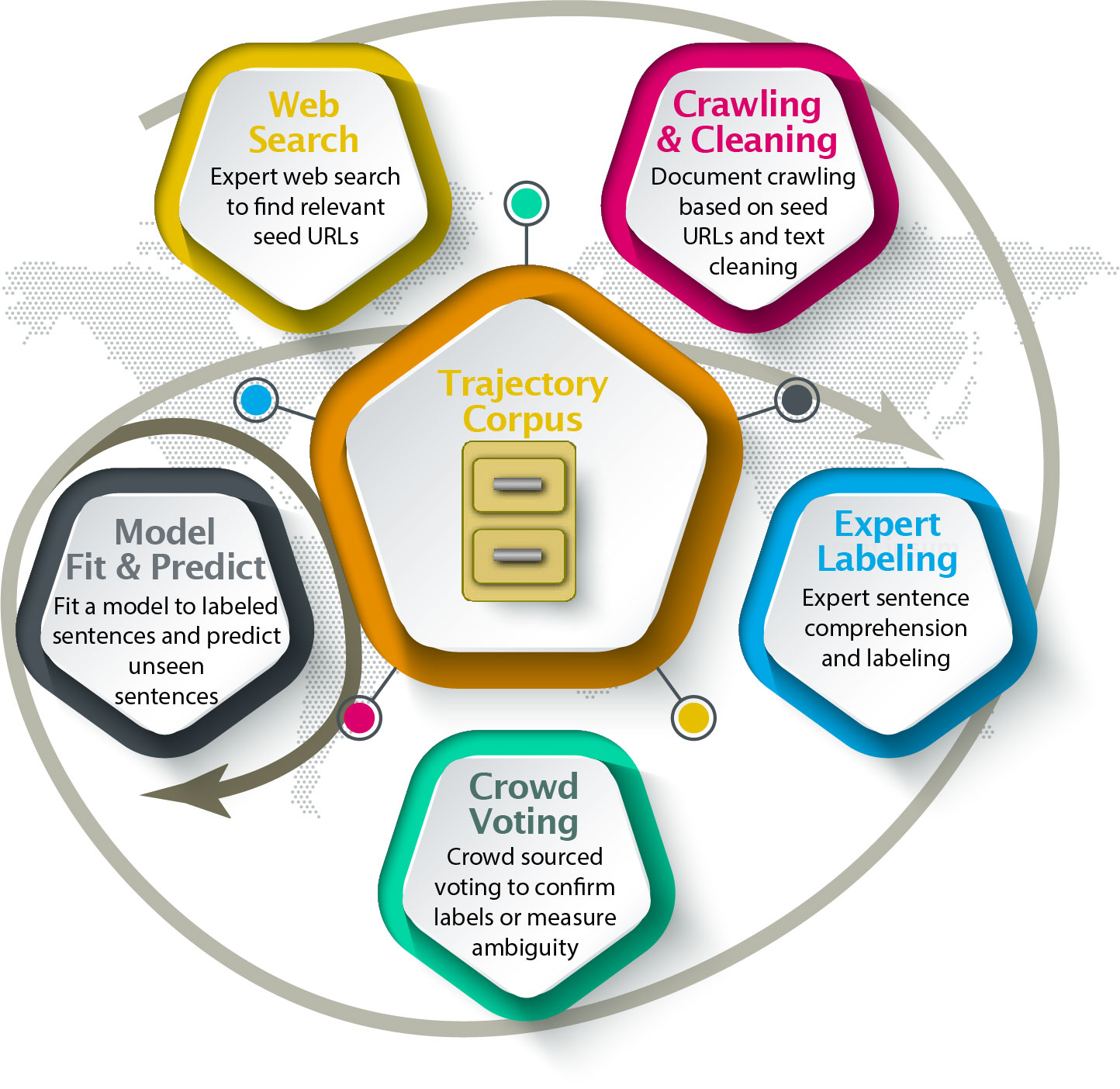} 
\caption{An iterative process for corpus creation from rarely occurring statements describing movement using humans and computers (shown in the lighter path) to the point where the final step of the model predicting labels can be repeated (shown in the darker path).}
\label{fig:workflow_diagram}
\end{center}
\end{figure}

As an additional contribution, we describe below our iterative process for creating the corpus which combines human labeling and machine learning and we show how it can produce a valuable corpus despite that the desired statements are relatively rare and difficult to find. We also make clear how similar corpora of rarely occurring text can be created using our process that starts with search and hand-labeling of a small seed training set, then uses machine learning models trained on that set that despite the small size can utilize modern advances in machine learning to produce more predicted samples for the corpus, and lastly uses human corrections of the predictions to quickly improve accuracy of the new samples found by the model (process shown in Figure~\ref{fig:workflow_diagram}). We discuss lessons learned from using multiple machine learning enhancements that greatly improve model predictions to address the cold start problem and in cases where only a small training set is available due to the prohibitive cost of generating very large training sets.

Therefore, the contributions of our work include 1) a corpus that can be used to a) improve detection of statements of geographic movement, which is currently underused information, b) improve GIR techniques by providing statements about geographic movement with many place mentions, non-geographic contextual information, and linguistic aspects describing geographic movement, c) aid spatial cognition researchers in understanding how people communicate and understand geographic movement; and 2) a method to create a corpus of rarely occurring text by bootstrapping human labeling efforts on a small seed set with machine learning predictions to produce a large corpus of high quality. Although other related efforts have improved the use of implicit geographic information in text with route directions \cite{jaiswal2012geocam}, historical exploration expeditions \cite{bekele2016historic}, routes \cite{drymonas2010routes}, hiking route description \cite{moncla2014hiking}, other paths \cite{moncla2014itinerary}, and geospatial natural language \cite{stock2013corpus}, etc., we believe no existing corpus is as large and diverse with respect to movement types.

\section{Related Work}

In this section, first, we review related work of geographic corpora that are most like ours. Next, because of the lack of comparable corpora, we discuss datasets that show similar information albeit with geographic and temporally precise features. Finally, we mention similar efforts to use human labeling as a starting point to allow machine learning predictions to comprise a corpus.

\subsection{Geographic Text Corpora}\label{sec:corpora}

GeoCorpora \cite{wallgrun2018geocorpora} is an example corpus
comprising of Twitter tweets where the geographic place mentions in the tweet text and user profiles were hand-coded identifying them as a place name and locating that place in a gazetteer. It does not have any movement explicitly tagged. This corpus is being used to improve GIR where place mentions are automatically detected in text and georeferenced to their correct location. GIR technique performance results can be compared with others used to experiment with other methods.

Another corpus that is comparable to our research is SpaceRef \cite{gotze2016spaceref}. SpaceRef is a corpus of approximately 1,400 statements describing landmarks along a walking route. The corpus is not a web corpus but was generated from people walking a route and then describing their route and landmarks used for navigating to the researchers to record as text. In addition to not being a web corpus, SpaceRef differs from our corpus because it focuses on landmarks across a small geographic area as opposed to geographically named features throughout the world. SpaceRef can be used to improve computational algorithms designed to assist in wayfinding and producing route descriptions. Like SpaceRef is the PURSUIT corpus \cite{blaylock2011annotation}, which also contains descriptions of routes and landmarks but from a car driving as compared to walking directions and with no effort to tie the landmark descriptions to the actual landmark geographically.

The Nottingham Corpus of Geospatial Language \cite{stock2013corpus} includes a broad range of uses of geospatial language in the form of sentences or clauses. Although it contains a substantial amount of statements of movement, it does not attempt to explicitly differentiate these from those that do not describe movement.

Lastly, the GeoCLEF activities provided text descriptions that include spatial language \cite{mandl2008geoclef}. This effort led to many advances in computational understanding of how people use spatial language with examples describing spatial movement, vague references to places, and relative spatial references, among others.

\subsection{Datasets of Geographic Movement Acquired from GPS Sensors}

Although text corpora with a geographic component are rare, many similar datasets exist that instead of containing text, contain explicit coordinates of location commonly acquired with accurate GPS. Although very valuable, these datasets ignore the implicit geography in text and our corpus can help fill this gap for researchers. The BrightKite check-in dataset and Gowalla check-in dataset are collections of social media check-ins that were used to show human travel patterns predicted by people's social networks \cite{cho2011friendship}. Additionally, Foursquare check-in datasets were created that all contain sparse spatial and temporal check-ins of users at locations where the location information is accurately input as latitude-longitude coordinates \cite{noulas2011empirical}. Other such datasets exist about human movement \cite{zheng2008geolife}, taxi trajectories \cite{moreira2012predictive}, bike sharing movement \cite{li2015traffic}, and animal movements \cite{rowland1997starkey}. These datasets have provided a valuable means to research human activity for urban planning and marketing, and to improve wildlife management. One more movement dataset to highlight is about hurricanes and allows for better prediction of future hurricane paths \cite{lee2007trajectory}. This sample of datasets are the more popular movement datasets, however, there are many more datasets available to allow for the study of various types of objects, people, animals, etc. moving through the World. Movebank is a repository of such datasets for trajectories and contains over one hundred datasets \cite{wikelski2019}.

In addition to the disadvantage that these datasets ignore information found in text, they lack the semantic and contextual information available in text that can allow people to better understand why the movement is occurring. Why did the movement start and end and what are the key decision points along the way are often readily available in text and a substantial body of research related to spatial cognition is focused on improving understanding of this geospatial information and deriving more knowledge from it \cite{richter2005context,klippel2005perception,klippel2009endpoint,klippel2005landmarks}. Therefore, text describing movement can supplement GPS datasets, be used to analyze movement on its own by applying GIR, and even provide contextual information that is more valuable for some applications than the geographic information.

\subsection{Computer Predicted Labels Merged with Human Labelers}

To address the challenges specific with creating corpora like ours, other efforts incorporate computer model predictions into the labeling process. The research described below use bootstrapping computer predictions to assist humans in labeling. Bootstrapping, in the context of labeling data, is an iterative process combining computer predicted labels with human labeler corrections \cite{pujara2011reducing}. An example of bootstrapping for labeled data is Inforex which is a web-based tool that allows for Named Entity Recognition (NER) and other NLP tools to identify important language features in text and have humans confirm these labels \cite{marcinczuk2012inforex}. Their tool is being used to find and annotate rare instances of text related to Polish suicide notes, Polish stock exchange notes, and other documents related to entities in the Polish language. While facing similar challenges, these researches took a similar approach to ours but with a subset of text readily available to them unlike ours where perhaps the biggest challenge is finding the samples. In a different project, a process like ours of using seed labels to train a model and create a corpus from predicted labels was used for a very different purpose with WordNets to create a sentiment corpus \cite{esuli2006sentiwordnet}. In later research on the same corpus, they also used a classifier by committee approach to improve predictions \cite{baccianella2010sentiwordnet}.

GeoCorpora (also mentioned in Section~\ref{sec:corpora}) is built in a similar fashion to ours in the sense of bootstrapping human labelers with computer predictions. Place names mentioned in text were initially identified and tied to a real-world place with automated GIR techniques and then humans either confirm or reject the labels based upon their contextual knowledge \cite{wallgrun2018geocorpora}. A web application was built that presented human labelers with a Tweet and asked them to confirm or correct the place mentions already identified by GIR, and add any missing place names.

\section{Corpus Building}

In order to create our gold-standard corpus of statements of geographic movement, we developed an iterative process (summarized in Figure~\ref{fig:workflow_diagram}) involving manual Web search for relevant seed URLs, harvesting of potentially relevant documents, expert human labeling, crowdsourced voting for agreement in labels, and training of a machine learning model from these labeled statements to predict more statements.

\subsection{Web Page Harvesting}

Since we are interested in a specific type of text, we first sought to subset potentially relevant documents on the Web. Therefore, we identified a small set of seed URLs through manual search which had content about animal migrations. National Geographic and eBird are two examples. The seed URLs were found using keyword searches such as \textit{bird migration} and \textit{human migration}. Scrapy web crawler (\url{https://scrapy.org/}) then followed links on these pages to download and store over 2.5 million potentially similar web pages in our PostgreSQL database. The BeautifulSoup (\url{https://www.crummy.com/software/BeautifulSoup/}) Python library extracted the main content from the web page and this was stored in a TSVector column type allowing for a full-text search. Lastly, GIR was applied on the text with GeoTxt \cite{karimzadeh2013geotxt} to identify place names. Using the text now in a database, search term queries related to animal migrations such as \textit{birds flying} and \textit{migrations} were performed to retrieve smaller batches. We added to our queries a parameter to only return documents that had multiple place mentions since these were more likely to describe geographic movement. For our first process iteration towards a seed labeled set, 17 initial documents were manually identified as having statements describing movement. Before taking this strategy of harvesting documents and filtering, we tried using commercial search engine queries to find relevant documents. This method did not produce good results as the web pages were commonly the site main pages without detailed content, sites quickly became irrelevant after the first page of results, and filtering for documents with multiple place mentions is not possible.

\subsection{Seed Set Human Labeling}

\begin{figure}[!h]
\begin{center}
\includegraphics[width=\columnwidth]{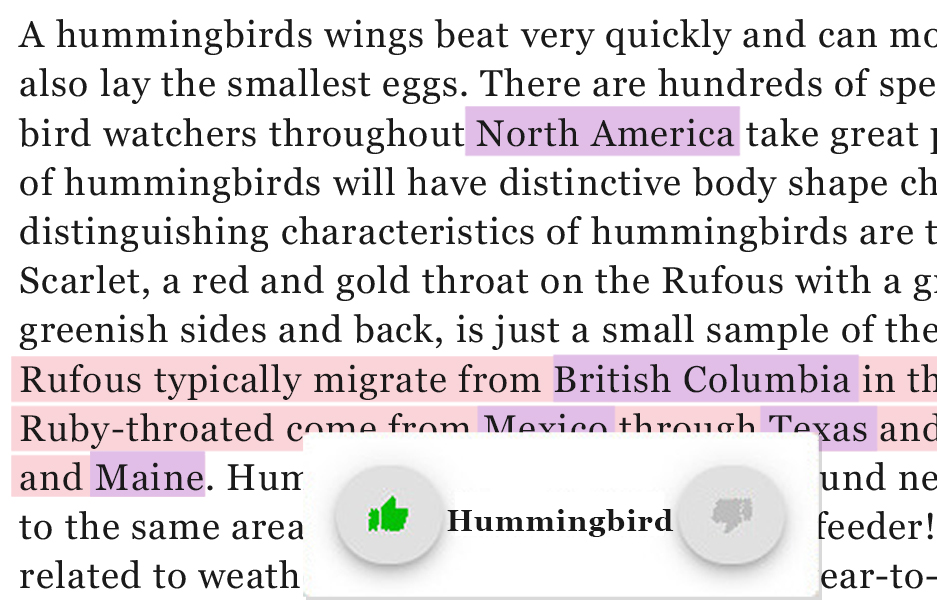} 
\caption{A portion of a document displayed in our labeling application with a statement of movement highlighted and a vote made to agree with the labels.}
\label{fig:labelerapp}
\end{center}
\end{figure}

Because of visual difficulties with reading the documents in either a text editor or spreadsheet, we developed a web application that presents the document text to the labeler, highlights the geographic locations in the text as hints to see text describing geographic movement, allows the labeler to highlight the desired statement and add the type of entity moving from a list of 471 animal, bird, fish, and other entity types, and stores the labels (by the start and end character indices for the highlighted text) to the database before automatically moving on to another document. Figure~\ref{fig:labelerapp} shows a portion of the labeling application.

It would likely be naive if we consider our own labels as truth given the ambiguous nature of text. Therefore, before training models with the labeled statements, Amazon Mechanical Turk (AMTurk) workers were employed to confirm the initial labels. This step was performed both to confirm the accuracy of the initial labels and to identify and measure ambiguity of each statement since text can often be interpreted differently. In order to maintain ethics, the AMTurk workers were paid per sentence at a rate that would equate to slightly more than the minimum hourly pay rate in the United States.

Our labeling application was modified slightly for voting and used to show the labels to AMTurk workers in order to either accept or reject the label. The AMTurk workers were asked to either agree with the label of movement and entity type, or disagree if either it is not describing movement or the entity type is incorrect. The workers were given some basic instructions and criteria for descriptions of geographic movement. The following sentence is an example where workers were not able to come to a majority of two-vote.

\begin{quotation}
 Even before the virus turned up in Turkey, the incidents at Qinghai and Erkhel and the spread of the H5N1 virus through Siberia and Kazakhstan had sparked new surveillance efforts \cite{normile426birds}.
\end{quotation}

The ambiguity in this sentence can be seen because first the movement is not explicit and second it would be questionable what is the entity moving, either the disease or the carriers of the disease. Therefore, is this sentence about birds moving or is it about the disease moving? And, what if a disease moves by infection and propagation and not via a bird? Should we consider that to be an instance of movement?

Each sentence was voted on by five different workers. Out of the 175 sentences labeled by us as describing movement, the AMTurk worker voting produced 124 agreements, four disagreements, and 47 sentences where the votes were undecided by a two-vote majority. The relatively high portion of sentences that were undecided illustrates the ambiguity in interpreting the text. Although this means 28\% of the sentences were undecided, we surmise that much less would be undecided if more workers voted on each document. Moreover, what is most important is not necessarily the final vote but that there does exist ambiguity in the sentence. We foresee users of the corpus utilizing the ambiguous labels differently towards their own goals, with perhaps some focusing primarily on analyzing the ambiguous sentences in order to find what makes them ambiguous. Since this hand labeling was time consuming, we sought computational ways to reduce production time of the corpus and to use these seed labels to predict more labels.

\subsection{Semi-Automated Corpus Expansion}

The hand labeled first set of statements was used to train machine learning models with the goal to predict more statements of movement thereby saving valuable time and costs related to manually finding more. Various models using vector counts, tf-idf words, tf-idf n-grams, tf-idf characters, or the cleaned sentence itself as features; and Fast Text \cite{joulin2017fasttext}, GloVe \cite{pennington2014glove}, or ELMo \cite{peters2018deep} as word embeddings were used. In total, 28 different model combinations were trained and evaluated on the corpus. Not only were the 28 implementations used to predict more samples, but this also gives a comparison on the performance of the various models against the corpus as baseline metrics for future users of the corpus to improve upon. These model results allowed us to later select the top performing models for improved predictions (see Section~\ref{sec:committee}).

\begin{table}[!h]
\begin{center}
\begin{tabularx}{\columnwidth}{|s|s|s|l|l|s|}
\hline
iter & Total in Set & Total after iter & TP & FP &  P  \\
\hline
0  & 4,718 & 4,718 & 175 & & \\ \hline
1   & 750  & 5,468     & 1     & 749          & 0.001            \\ \hline 
2   & 832  & 6,300   & 1     & 831          & 0.001           \\ \hline
3   & 796  & 7,096   & 1     & 795          & 0.001           \\ \hline 
4   & 790  & 7,886  & 0     & 790          & 0.0             \\ \hline
5   & 809  & 8,695  & 37     & 772          & 0.05            \\ \hline
6   & 661  & 9,356  & 24     & 637          & 0.04            \\ \hline
7   & 646  & 10,002  & 77     & 569          & 0.12           \\ \hline
8   & 696  & 10,698  & 107    & 589          & 0.15           \\ \hline
9   & 644  & 11,342  & 95     & 549          & 0.15           \\ \hline
10  & 411  & 11,753  & 105    & 306          & \textbf{0.25}   \\ \hline
\end{tabularx}
\caption{Iteration results show an increase in the number of true positives (TP) and precision (P) as more labels and training data are added.}\label{tab:iteration_acc}
\end{center}
\end{table}

Scikit-Learn was used for Logistic Regression, Random Forest, and Support Vector Machine (SVM) models \cite{pedregosa2011scikitlearn} while the Python implementation (\url{https://xgboost.readthedocs.io}) of XGBoost \cite{chen2016xgboost} was used. The various deep learning models (as well as vector counts and tf-idf calculations) were recommended by \newcite{shivan2019bansal} because of their success in text classification and the use of ELMo Embeddings in a simple feed-forward neural network was illustrated by \newcite{mahapatra2019elmokeras}. Since the positive samples were far less than the negative samples, the positives were oversampled using the SMOTE \cite{chawla2002smote} method or random oversampling where SMOTE was not possible. Lastly, the text was cleaned of punctuation, uncommon characters, and contractions. The case was not converted to lowercase but left as-is since geographic place mentions are assumed to be important to descriptions of movement and are commonly capitalized.

\begin{table}[!h]
\begin{center}
\begin{tabularx}{\columnwidth}{|a|y|y|}
\hline
model  & Precision  & F-Measure \\
\hline
Simple Neural Network and ELMo  & \textbf{0.42}     & \textbf{0.69} \\
\hline
Random Forest and tf-idf\_ngrams   &  0.33     & 0.67   \\ 
\hline
XGBoost and tf-idf\_ngrams   &  0.31     & 0.65    \\   
\hline
Logistic Regression and tf-idf\_ngrams   &  0.27     & 0.65    \\ 
\hline
SVM and tf-idf\_ngrams  &  0.26     & 0.65    \\ 
\hline
CNN with FastText    &  0.22     & 0.61    \\      
\hline
\end{tabularx}
\caption{Precision and F-Measure for selected top performing models. Non-deep models in general were better performing with the exception of a simple neural network with ELMo embeddings being the best.}
\label{tab:topmodels}
\end{center}
\end{table}

\begin{table}[!h]
\begin{center}
\begin{tabularx}{\columnwidth}{|X|}
\hline
North of the Qinghai-Tibetan Plateau, birds from both populations traveled largely without stops across the Gobi Desert en route to and from breeding grounds in Mongolia \cite{Palm2015}. \\
\hline
Made the flagship of an American squadron, led by Commodore Richard Morris, Chesapeake sailed for the Mediterranean in April and arrived at Gibraltar on May 25 \cite{hickman2020chesapeake}. \\
\hline
Traffic was not helping but within an hour I was on the F3 freeway heading north towards Gosford \cite{deguara2002storms}. \\
\hline
\color{darkgray}{This model effectively quantifies the relative importance of different migration corridors and stopover sites and may help prioritize specific areas in flyways for conservation of waterbird populations \cite{Palm2015}.} \\
\hline
\color{darkgray}{Other migrating birds rarely attempt to keep such a precise schedule \cite{perkypet2020birds}.} \\
\hline
\color{darkgray}{While influenza virus surveillance in Alaskan waterfowl species found predominantly LPAIV (15), frequent reassortment was noted in one study of northern pintails, with nearly half (44.7\%) of the LPAIV tested having at least one gene segment demonstrating closer relatedness to Eurasian than to North American LPAIV genes (16) \cite{lee2015spread}.} \\
\hline
\end{tabularx}
\caption{Examples of true positive (TP) statements of movement are shown in black colored text while examples of false positive (FP) statements that do not describe movement are shown in gray colored text.}
\label{tab:examplestatements}
 \end{center}
\end{table}

In each iteration, all models were trained and evaluated on the corpus using an 80/20 training and testing split. Next, the top five performing models were trained on the entire corpus and then used to predict more statements of movement on unseen documents from the initial harvested ones. The top performing models can be seen in Table~\ref{tab:topmodels}. Again, as seen from the lighter path in Figure~\ref{fig:workflow_diagram}, the highest probability positively predicted statements are evaluated by hand by the expert and crowd to confirm the labels. Results of the predictions as the iterations were completed show an increase in the models' performances (as seen in Table~\ref{tab:iteration_acc}). As an example of the procedure, in iteration three the models were trained on the 6,300 statements that were the total after iteration two. From these models, 796 new statements combined were predicted as positive by the top five models, and out of these 796 positives, only one was a true positive from the hand evaluation. The corrected 796 statements are added back to the total after iteration three making 7,096 for the next iteration. Although the precision of the predictions increased with more training data, the highest precision was only 0.25; so we sought improvements to the models. An interesting conclusion from this iterative labeling is despite that in early iterations very few positive labels were identified and added to the training data, the precision began steadily increasing. This suggests that even providing more negative samples for the classifiers increased learning on what does not describe movement thereby improving positive predictions. At the conclusion of these ten iterations where humans evaluated the predictions, and including the initial sets, the final gold-standard corpus contained 11,753 statements with 623 of these describing movement. The statements consist of 30 different types of entities moving from multiple different types of birds, animals, and fish, humans using different modes of transportation, to goods being shipped.

Table~\ref{tab:examplestatements} shows three true positive statements that describe movement and three false positive statements where the models predicted movement incorrectly. It is clear in the false positives that given the small amount of training data and simple models, the models focused on place mentions and entity types but did not understand some of the linguistic clues that differentiate movement from descriptions where the entities are simply at locations. We looked to address this issue through model enhancements, specifically by using word embeddings (Section~\ref{sec:embeddings}) and ensembling models (Section~\ref{sec:committee}).

\section{Model Enhancements Towards Automated Expansion}

Two model enhancements greatly improved predictions which allowed us to progress from difficult to find statements describing movement to highly accurate predicted statements. As we can see from Table~\ref{tab:topmodels}, the non-deep models consistently performed well with four of the top five models. However, using ELMo word embeddings in the simple deep model performed the best. This prompted the question: \textit{Is the deep model performing the best because of the model itself or is it because of the ELMo embeddings used?} To answer this question, we added word embeddings for the non-deep models. Secondly, since we were already using multiple models for predictions, we decided to test using a classifier by committee approach to predictions.

\subsection{Word Embeddings}\label{sec:embeddings}

We used the same non-deep models as earlier, but before training the model we added ELMo embeddings to the text using the pre-trained ELMo embeddings version 2 available on TensorFlow HUB \url{https://tfhub.dev/google/elmo/2}. Embeddings were calculated on a sentence-by-sentence basis with a maximum of 100 words per sentence to save computational resources and the resulting vectors were used to train the models (as described in \newcite{joshi2019elmo}). Table~\ref{tab:elmo_non-deep} shows that all four of the non-deep models performed better with the ELMo embeddings as compared to without in each of the two metrics shown. Our results that ELMo embeddings improved the non-deep models is consistent with those presented in \newcite{Maslennikova2019ELMo} where ELMo improved SVM for classifying news articles. The non-deep models performing better than the deep models is likely because our amount of labeled data is relatively small. Therefore, using word embeddings with non-deep models shows promise for other machine learning prediction tasks on text when it is not feasible to have a large amount of training data. This model enhancement of using ELMo embeddings in non-deep models showed promise that we can use the predictions for our silver-standard corpus, but better predictions were still required.

\begin{table}[!h]
\begin{center}
\begin{tabularx}{\columnwidth}{|y|y|y|y|y|}
\hline
 model & \textcolor{black}{Precision w/ ELMo} & \textcolor{gray}{Precision w/o ELMo} & \textcolor{black}{F-Measure w/ ELMo} & \textcolor{gray}{F-Measure w/o ELMo} \\
\hline
Random \newline Forest      & \textcolor{black}{0.59}       & \textcolor{gray}{0.33}        & \textcolor{black}{0.69}           & \textcolor{gray}{0.67}              
\\ \hline
SVM                & \textcolor{black}{0.38}      & \textcolor{gray}{0.26}        & \textcolor{black}{0.72}          & \textcolor{gray}{0.65}                   \\
\hline
LogReg              & \textcolor{black}{0.37}      & \textcolor{gray}{0.27}        & \textcolor{black}{0.73}          & \textcolor{gray}{0.65}               \\
\hline
XGBoost            & \textcolor{black}{0.37}      & \textcolor{gray}{0.31}        & \textcolor{black}{0.73}          & \textcolor{gray}{0.65}                  \\
\hline
\end{tabularx}
\caption{Metrics show that predictions with (w/) ELMo Embeddings (in black colored text) improve in all four non-deep models for statements of movement compared to without (w/o) ELMo Embeddings (in grey colored text).}
\label{tab:elmo_non-deep}
 \end{center}
\end{table}

\subsection{Classifier by Committee}\label{sec:committee}

Our second effort to improve results was to use a simple ensemble method to combine the top performing model predictions. Since our focus is on predicting statements of movement correctly, we felt that taking the predictions where the top performing models agreed on the positive prediction would produce better results. Therefore, we again used our existing corpus and trained and tested all our models. Next, we tested multiple simple ways to ensemble the top performing models, including a maximum vote where the class with the highest number of votes between the five models is chosen and a second where the mean probability between the five models is taken. We sorted the models by their F-Scores to select the top five best performing models. The results of this experiment can be seen in Table~\ref{tab:ensemble}. The ensemble methods of mean probability and maximum vote showed similar substantial improvements as compared to the best performing individual model, which had already been improved with ELMo Embeddings. We use precision as a metric since we are more interested in predicting positive statements while accepting that some positives are missed as false negatives.

Given the success of the ensemble models, we decided to use the method of the top five performing models, as sorted by the F-Score (these models are: Random Forest with ELMo, SVM with ELMo, Logistic Regression with ELMo, XGBoost with ELMo, and Simple Neural Network with ELMo). We trained these models on our entire gold-standard corpus and then made predictions on unseen sentences in the same way as our process iterations before. We then used the ensemble method, selected the 100 sentences with the highest probability, and hand-evaluated these. In our previous iterations, the Simple Neural Network with ELMo embeddings produced the highest portion of true positives in a similar experiment. For this top performing individual model, 81 of the 100 sentences with the highest probabilities were true positives. For the ensemble method, all 100 of the highest probability sentences were true positives. Since all 100 of the sentences described movement, we then hand-evaluated more than the 100 sentences with the highest probabilities. The sentences with the highest probabilities were all true positives until the 400\textsuperscript{th} statement. This first false positive statement had an average probability of 0.77 from the ensemble model prediction.

\begin{table}[!h]
\begin{center}
\begin{tabularx}{\columnwidth}{|b|s|s|}
\hline
    & Precision & F-score  \\
\hline
Mean Probability - Top 5 Sorted by F-Measure & \textbf{0.91} & \textbf{0.88} \\ \hline
Max Vote - Top 5 Sorted by F-Measure         & 0.91 & 0.87 \\ \hline
Random Forest with ELMo           & 0.59 & 0.69 \\ \hline
\end{tabularx}
\caption{Ensembling the top performing models produced better metrics as compared to a selected top performing model by itself.}
\label{tab:ensemble}
 \end{center}
\end{table}

By using ELMo embeddings in the top performing models which were non-deep models and by taking a classifier by committee ensemble of the top performing models, we were able to improve predictions for descriptions of movement to the point that model predictions can now be used to create the statements of movement in our large silver-standard corpus with confidence that the error rate will be very low. Negative samples for the silver-standard corpus are randomly chosen from the harvested documents that are confidently negative to match the number of positive statements to have a balanced distribution. Both the average probability of predictions and the votes of the models as measures of the likelihood of error in each statement are included in the corpus. We foresee users of the corpus being able to correct errors in the corpus while they use it.

\section{Corpus Use-Case}

\begin{figure}[!h]
\begin{center}
\includegraphics[width=\columnwidth]{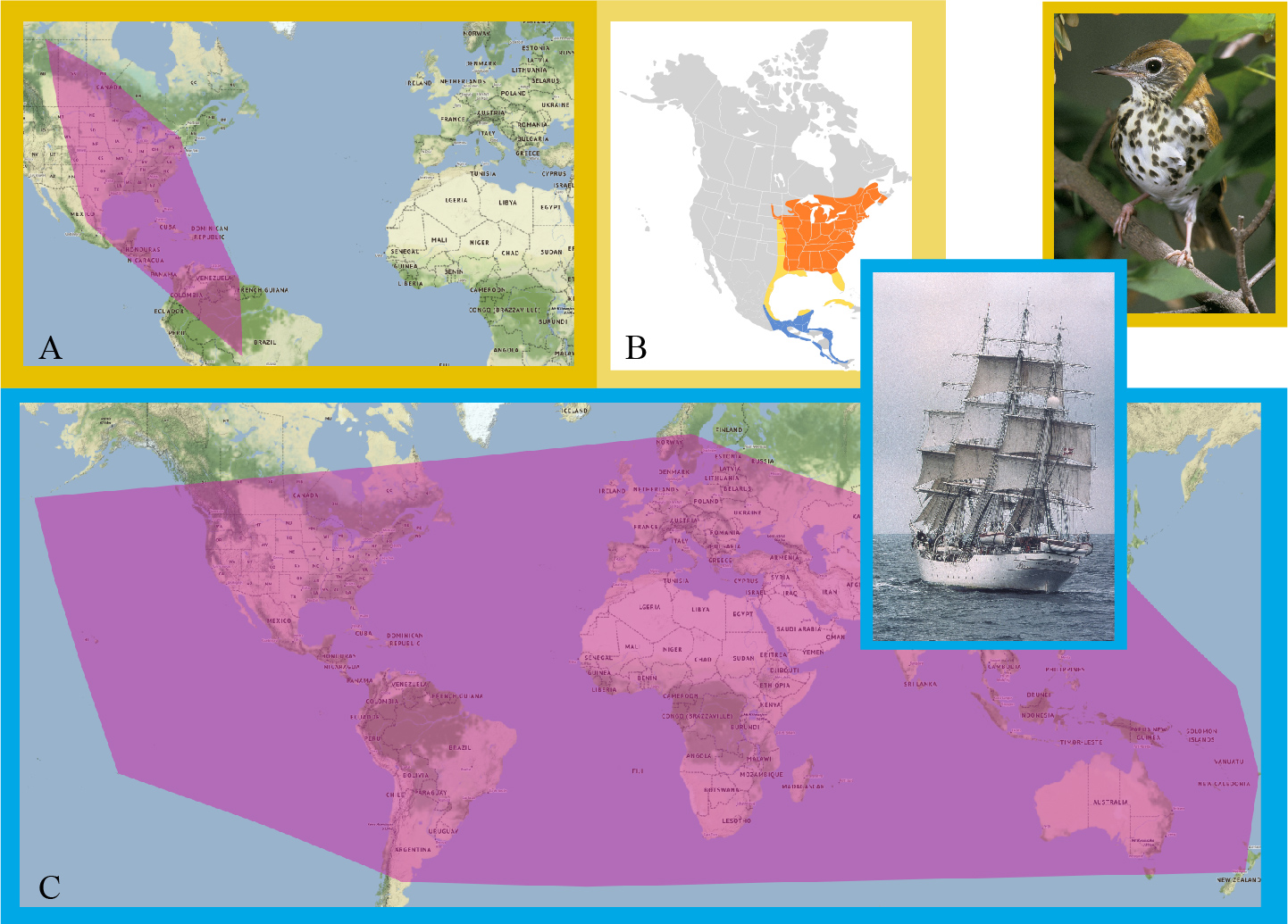} 
\caption{Wood Thrush birds show a very different range by mapping the place mentions (A) in their movement statements as compared to the movement range of ships similarly mapped (C). The Wood Thrush range mapped from our text corpus (A) roughly corresponds to its documented actual range (B) \protect\cite{evans2011thrushmap}.}
\label{fig:usecase}
\end{center}
\end{figure}

Our corpus is designed to help other researchers improve techniques to detect statements of movement in text. With improved techniques to detect statements of movement, the geographic information and other information described can be better utilized. Figure~\ref{fig:usecase} shows an example use-case of our corpus where the extent of all statements of movement from two different entity types, wood thrush birds (Figure~\ref{fig:usecase}A) and ships (Figure~\ref{fig:usecase}C), are mapped using GIR to convert the text place mentions to their real-world locations. Not surprisingly, the statements of movement of the Wood Thrush mapped with GIR show a range primarily covering North America and into South America. This range is consistent with the documented migration range of the Wood Thrush (Figure~\ref{fig:usecase}B). The variation between the mapped range and the documented range in Canada and Brazil is only because the GIR software places these countries mentions in the center of each country on the map while the birds' range likely extends only slightly into the southern and northern areas, respectively. In contrast to the range of the Wood Thrush are the statements of movement from ships, which are mainly from descriptions of exploration voyages in the Age of Discovery. Once again, the overall pattern of the movement mapped is what you would expect with a nearly global coverage and covering the oceans.

This use-case is a comparison against already known ranges in order to show potential. It is a simple use of the corpus in mapping implicit geography that is found in statements of movement in text. Examples where the movement is not already mapped could be historical journals of migrations, other historical movement, and illegal trafficking of animals, goods, or humans where the information is either text or voice communication that can be converted to text (assuming such statements are added to the corpus). There are many other potential uses for the corpus beyond this example like efforts to improve GIR with context or research in the field of Spatial Cognition to understand how people communicate movement.

\section{Conclusions and Future Work}

Towards the goal of creating a corpus of statements of geographic movement from web documents, three overall challenges exist. First, statements of geographic movement on the Web are relatively few and difficult to find in proportion to the size of the Web. Second, reading, understanding, and labeling such text is time-consuming. Third, ambiguity exists in the text and therefore identifying which are statements of movement and which are not is a difficult problem. To solve these challenges, an iterative approach was developed involving initial human manual search and labeling, computer model predictions of more statements, and humans confirming the model predictions. Our simple corpus creation process shows how manual, crowdsourced, computational, and visual techniques can be merged to produce high-quality training data of rare-event samples. Our model predictions improved as process iterations increased and enhanced model techniques show that our process can now produce a large silver-standard version of the corpus where a small amount of error is accepted but the time-consuming human labeling task is removed.

Our initial work produced a gold-standard corpus of statements of movement. While creating the gold-standard corpus, results of the machine learning on the initial small seed set show that non-deep models were more consistent on the small seed subset. Also, word embeddings improved the non-deep model's performances and ensembling models further improved predictions to the point that they can be used to enlarge the silver-standard corpus without further human effort. Although our corpus is intended to be broad in its coverage, future potential target entity types include statements of movement of illegally trafficked goods, firearms, animals, and humans with the goal of helping detection of such statements.

\begin{figure}[!h]
\begin{center}
\includegraphics[width=\columnwidth]{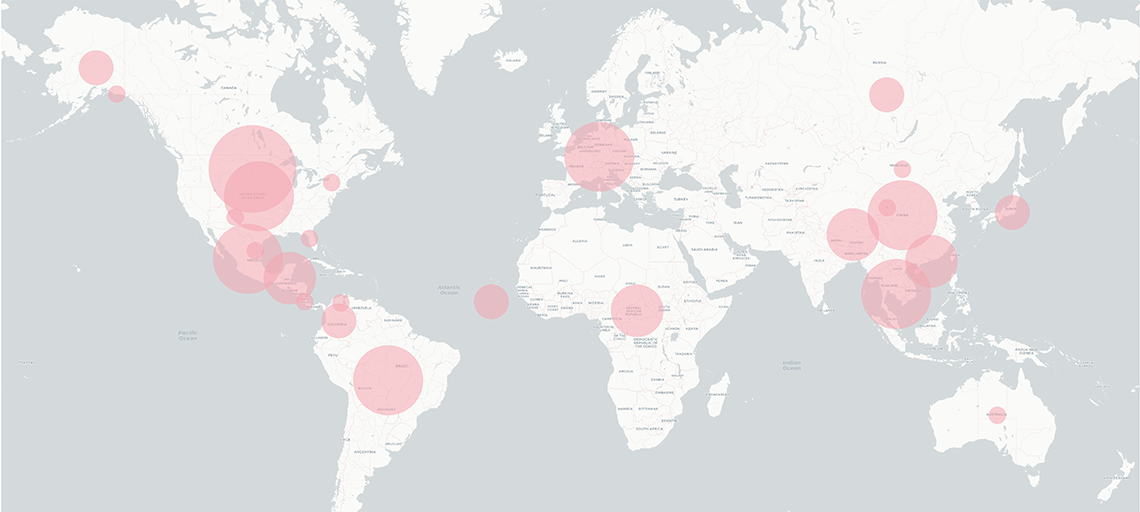}
\caption{All place mentions in the corpus mapped shows the global distribution and accessibility of the corpus.}
\label{fig:corpusplaces}
\end{center}
\end{figure}

Our substantial manual and computational efforts will allow other researchers to benefit by spurring more research towards utilizing the implicit geography in statements of movement and better understand how people conceptualize geographic movement. We give one use-case example of the corpus and more potential uses are described. When all place mentions in the corpus are mapped with GIR, the distribution is diverse globally (shown in Figure~\ref{fig:corpusplaces}) which should also make the corpus more useful to a wide audience.

Although we can already extend our corpus automatically with statements of movement, we have not yet automatically labeled the type of entity moving. As an experiment towards this future work, we used spaCy 2 Python toolkit's Part-of-Speech (POS) Tagger (\url{https://spacy.io/usage/linguistic-features}) on the 20 statements with the highest probability predictions. Out of these 20 statements, 11 of the actual entity types as manually found were labeled by the tagger as nominal subjects in the sentence, five were the object of the preposition, and four were the root or a compound root. Since a simple POS Tagger showed consistency in the entity type being only a few specific parts-of-speech, we anticipate being able to identify the entity types in our future work using a POS Tagger and a rules-based approach.

\section{Acknowledgements}

We kindly thank Dr. Alan M. MacEachren for his suggestions to improve the corpus and to make the corpus more valuable to a larger audience.

\section{Bibliographical References}\label{reference}

\bibliographystyle{lrec}
\bibliography{PezanowskiMitra_GeographicMovementInText}

\begin{thebibliography}{}

\bibitem[\protect\citename{Baccianella \bgroup et al.\egroup
  }2010]{baccianella2010sentiwordnet}
Baccianella, S., Esuli, A., and Sebastiani, F.
\newblock (2010).
\newblock {SentiWordNet 3.0: An Enhanced Lexical Resource for Sentiment
  Analysis and Opinion Mining}.
\newblock In Nicoletta Calzolari~(Conference Chair), et~al., editors, {\em
  {Proceedings of the Seventh International Conference on Language Resources
  and Evaluation (LREC'10)}}, pages 2200--2204, Valletta, Malta. European
  Language Resources Association (ELRA).

\bibitem[\protect\citename{Bansal}2018]{shivan2019bansal}
Bansal, S.
\newblock (2018).
\newblock {A Comprehensive Guide to Understand and Implement Text
  Classification in Python}.

\bibitem[\protect\citename{Bekele \bgroup et al.\egroup
  }2016]{bekele2016historic}
Bekele, M.~K., De~By, R.~A., and Singh, G.
\newblock (2016).
\newblock {Spatiotemporal Information Extraction from a Historic Expedition
  Gazetteer}.
\newblock {\em {ISPRS International Journal of Geo-Information}}, 5(12).

\bibitem[\protect\citename{Blaylock}2011]{blaylock2011annotation}
Blaylock, N.
\newblock (2011).
\newblock {Semantic Annotation of Street-Level Geospatial Entities}.
\newblock In {\em {2011 IEEE Fifth International Conference on Semantic
  Computing}}, pages 444--448, Sep.

\bibitem[\protect\citename{Chawla \bgroup et al.\egroup }2002]{chawla2002smote}
Chawla, N.~V., Bowyer, K.~W., Hall, L.~O., and Kegelmeyer, W.~P.
\newblock (2002).
\newblock {SMOTE: Synthetic Minority over-Sampling Technique}.
\newblock {\em Journal of Artificial Intelligence Research}, 16(1):321–357,
  Jun.

\bibitem[\protect\citename{Chen and Guestrin}2016]{chen2016xgboost}
Chen, T. and Guestrin, C.
\newblock (2016).
\newblock {XGBoost: A Scalable Tree Boosting System}.
\newblock In {\em {Proceedings of the 22nd ACM SIGKDD International Conference
  on Knowledge Discovery and Data Mining}}, KDD '16, pages 785--794, New York,
  NY, USA. ACM.

\bibitem[\protect\citename{Cho \bgroup et al.\egroup }2011]{cho2011friendship}
Cho, E., Myers, S.~A., and Leskovec, J.
\newblock (2011).
\newblock {Friendship and Mobility: User Movement in Location-Based Social
  Networks}.
\newblock In {\em {Proceedings of the 17th ACM SIGKDD International Conference
  on Knowledge Discovery and Data Mining}}, KDD ’11, page 1082–1090, New
  York, NY, USA. Association for Computing Machinery.

\bibitem[\protect\citename{Deguara}2002]{deguara2002storms}
Deguara, J.
\newblock (2002).
\newblock {Supercell Outbreak - 8th February 2002 - including video footage}.

\bibitem[\protect\citename{Dodge \bgroup et al.\egroup }2012]{Dodge2012}
Dodge, S., Laube, P., and Weibel, R.
\newblock (2012).
\newblock {Movement similarity assessment using symbolic representation of
  trajectories}.
\newblock {\em International Journal of Geographical Information Science},
  26(9):1563--1588.

\bibitem[\protect\citename{Dodge \bgroup et al.\egroup }2016]{Dodge2016a}
Dodge, S., Weibel, R., Ahearn, S.~C., Buchin, M., and Miller, J.~A.
\newblock (2016).
\newblock {Analysis of movement data}.
\newblock {\em International Journal of Geographical Information Science},
  30(5):825--834.

\bibitem[\protect\citename{Dodge}2016a]{Dodge2016b}
Dodge, S.
\newblock (2016a).
\newblock Context-sensitive spatiotemporal simulation model for movement.
\newblock In {\em {International Conference on GIScience Short Paper
  Proceedings}}, volume~1, pages 80--83, Montreal, Canada.

\bibitem[\protect\citename{Dodge}2016b]{Dodge2016}
Dodge, S.
\newblock (2016b).
\newblock {From Observation to Prediction: The Trajectory of Movement Research
  in GIScience}.
\newblock {\em Advancing Geographic Information Science: The Past and Next
  Twenty Years}, pages 123 -- 136.

\bibitem[\protect\citename{Drymonas and Pfoser}2010]{drymonas2010routes}
Drymonas, E. and Pfoser, D.
\newblock (2010).
\newblock {Geospatial Route Extraction from Texts}.
\newblock In {\em {Proceedings of the 1st ACM SIGSPATIAL International Workshop
  on Data Mining for Geoinformatics}}, DMG '10, pages 29--37, New York, NY,
  USA. ACM.

\bibitem[\protect\citename{Esuli and Sebastiani}2006]{esuli2006sentiwordnet}
Esuli, A. and Sebastiani, F.
\newblock (2006).
\newblock {SentiWordNet: A Publicly Available Lexical Resource for Opinion
  Mining}.
\newblock In {\em {Proceedings of the Fifth International Conference on
  Language Resources and Evaluation (LREC'06)}}, pages 417--422, Genoa, Italy.
  European Language Resources Association (ELRA).

\bibitem[\protect\citename{{Evans, M., Gow, E., Roth, R. R., Johnson, M. S.,
  and Underwood, T. J.}}2011]{evans2011thrushmap}
{Evans, M., Gow, E., Roth, R. R., Johnson, M. S., and Underwood, T. J.}
\newblock (2011).
\newblock {Wood Thrush (Hylocichla mustelina), version 2.0}.
\newblock In A.~F. Poole, editor, {\em {The Birds of North America}}, pages
  201--213. Cornell Lab of Ornithology, Ithaca, NY, USA.

\bibitem[\protect\citename{Gonz{\'{a}}lez \bgroup et al.\egroup
  }2008]{Gonzalez2008}
Gonz{\'{a}}lez, M.~C., Hidalgo, C.~A., and Barab{\'{a}}si, A.-L.
\newblock (2008).
\newblock {Understanding individual human mobility patterns}.
\newblock {\em Nature}, 453(7196):779--782.

\bibitem[\protect\citename{G{\"o}tze and Boye}2016]{gotze2016spaceref}
G{\"o}tze, J. and Boye, J.
\newblock (2016).
\newblock {SpaceRef: A corpus of street-level geographic descriptions}.
\newblock In {\em {Proceedings of the Tenth International Conference on
  Language Resources and Evaluation (LREC'16)}}, pages 3822--3827.

\bibitem[\protect\citename{Hickman}2020]{hickman2020chesapeake}
Hickman, K.
\newblock (2020).
\newblock {War of 1812: USS Chesapeake}.

\bibitem[\protect\citename{Huang}2017]{Huang2017}
Huang, Q.
\newblock (2017).
\newblock {Mining online footprints to predict user’s next location}.
\newblock {\em International Journal of Geographical Information Science},
  31(3):523--541.

\bibitem[\protect\citename{Jaiswal \bgroup et al.\egroup
  }2010]{jaiswal2012geocam}
Jaiswal, A., Pezanowski, S., Mitra, P., Zhang, X., Xu, S., Turton, I., Klippel,
  A., and MacEachren, A.~M.
\newblock (2010).
\newblock {GeoCAM: A geovisual analytics workspace to contextualize and
  interpret statements about movement}.
\newblock {\em Journal of Spatial Information Science}, pages 279--294.

\bibitem[\protect\citename{Joshi}2019]{joshi2019elmo}
Joshi, P.
\newblock (2019).
\newblock {A Step-by-Step NLP Guide to Learn ELMo for Extracting Features from
  Text}.

\bibitem[\protect\citename{Joulin \bgroup et al.\egroup
  }2017]{joulin2017fasttext}
Joulin, A., Grave, E., Bojanowski, P., and Mikolov, T.
\newblock (2017).
\newblock {Bag of Tricks for Efficient Text Classification}.
\newblock In {\em {Proceedings of the 15th Conference of the European Chapter
  of the Association for Computational Linguistics: Volume 2, Short Papers}},
  pages 427--431. Association for Computational Linguistics, Apr.

\bibitem[\protect\citename{Ju \bgroup et al.\egroup }2016]{ju2016things}
Ju, Y., Adams, B., Janowicz, K., Hu, Y., Yan, B., and McKenzie, G.
\newblock (2016).
\newblock {Things and Strings: Improving Place Name Disambiguation from Short
  Texts by Combining Entity Co-Occurrence with Topic Modeling}.
\newblock In Eva Blomqvist, et~al., editors, {\em {Knowledge Engineering and
  Knowledge Management}}, pages 353--367, Cham. Springer International
  Publishing.

\bibitem[\protect\citename{Karimzadeh \bgroup et al.\egroup
  }2013]{karimzadeh2013geotxt}
Karimzadeh, M., Huang, W., Banerjee, S., Wallgr\"{u}n, J.~O., Hardisty, F.,
  Pezanowski, S., Mitra, P., and MacEachren, A.~M.
\newblock (2013).
\newblock {GeoTxt: A Web API to Leverage Place References in Text}.
\newblock In {\em {Proceedings of the 7th Workshop on Geographic Information
  Retrieval}}, GIR ’13, page 72–73, New York, NY, USA. Association for
  Computing Machinery.

\bibitem[\protect\citename{Klippel and Li}2009]{klippel2009endpoint}
Klippel, A. and Li, R.
\newblock (2009).
\newblock {The Endpoint Hypothesis: A Topological-Cognitive Assessment of
  Geographic Scale Movement Patterns}.
\newblock In Kathleen~Stewart Hornsby, et~al., editors, {\em {Spatial
  Information Theory}}, pages 177--194, Berlin, Heidelberg. Springer Berlin
  Heidelberg.

\bibitem[\protect\citename{Klippel and Winter}2005]{klippel2005landmarks}
Klippel, A. and Winter, S.
\newblock (2005).
\newblock {Structural Salience of Landmarks for Route Directions}.
\newblock In Anthony~G. Cohn et~al., editors, {\em {Spatial Information
  Theory}}, pages 347--362, Berlin, Heidelberg. Springer Berlin Heidelberg.

\bibitem[\protect\citename{Klippel \bgroup et al.\egroup
  }2005]{klippel2005perception}
Klippel, A., Tappe, H., Kulik, L., and Lee, P.~U.
\newblock (2005).
\newblock {Wayfinding choremes—a language for modeling conceptual route
  knowledge}.
\newblock {\em Journal of Visual Languages \& Computing}, 16(4):311 -- 329.
\newblock Perception and ontologies in visual, virtual and geographic space.

\bibitem[\protect\citename{Lee \bgroup et al.\egroup }2007]{lee2007trajectory}
Lee, J.-G., Han, J., and Whang, K.-Y.
\newblock (2007).
\newblock {Trajectory Clustering: A Partition-and-Group Framework}.
\newblock In {\em {Proceedings of the 2007 ACM SIGMOD International Conference
  on Management of Data}}, SIGMOD ’07, page 593–604, New York, NY, USA.
  Association for Computing Machinery.

\bibitem[\protect\citename{Lee \bgroup et al.\egroup }2015]{lee2015spread}
Lee, D.-H., Torchetti, M.~K., Winker, K., Ip, H.~S., Song, C.-S., and Swayne,
  D.~E.
\newblock (2015).
\newblock {Intercontinental Spread of Asian-Origin H5N8 to North America
  through Beringia by Migratory Birds}.
\newblock {\em Journal of Virology}, 89(12):6521--6524.

\bibitem[\protect\citename{Li \bgroup et al.\egroup }2015]{li2015traffic}
Li, Y., Zheng, Y., Zhang, H., and Chen, L.
\newblock (2015).
\newblock {Traffic Prediction in a Bike-Sharing System}.
\newblock In {\em {Proceedings of the 23rd SIGSPATIAL International Conference
  on Advances in Geographic Information Systems}}, SIGSPATIAL ’15, New York,
  NY, USA. Association for Computing Machinery.

\bibitem[\protect\citename{MacEachren}2014]{maceachren2014place}
MacEachren, A.
\newblock (2014).
\newblock {Place Reference in Text as a Radial Category: A Challenge to Spatial
  Search, Retrieval, and Geographical Information Extraction from Documents
  that Contain References to Places}.
\newblock In {\em {2014 Specialist Meeting — Spatial Search}}, Santa Barbara,
  CA. UCSB Center for Spatial Studies.

\bibitem[\protect\citename{Mahapatra}2019]{mahapatra2019elmokeras}
Mahapatra, S.
\newblock (2019).
\newblock {Transfer Learning using ELMO Embeddings}.

\bibitem[\protect\citename{Mandl \bgroup et al.\egroup }2009]{mandl2008geoclef}
Mandl, T., Carvalho, P., Di~Nunzio, G.~M., Gey, F., Larson, R.~R., Santos, D.,
  and Womser-Hacker, C.
\newblock (2009).
\newblock {GeoCLEF 2008: The CLEF 2008 Cross-Language Geographic Information
  Retrieval Track Overview}.
\newblock In Carol Peters, et~al., editors, {\em {Evaluating Systems for
  Multilingual and Multimodal Information Access}}, pages 808--821, Berlin,
  Heidelberg. Springer Berlin Heidelberg.

\bibitem[\protect\citename{Marcinczuk \bgroup et al.\egroup
  }2012]{marcinczuk2012inforex}
Marcinczuk, M., Kocon, J., Broda, B., {Marci{\'{n}}czuk Micha{\l}and
  Koco{\'{n}}}, J., and Broda, B.
\newblock (2012).
\newblock {Inforex-a web-based tool for text corpus management and semantic
  annotation}.
\newblock In {\em {Proceedings of the Eighth International Conference on
  Language Resources and Evaluation (LREC'12)}}, pages 224--230, Istanbul,
  Turkey. European Language Resources Association (ELRA).

\bibitem[\protect\citename{Maslennikova}2019]{Maslennikova2019ELMo}
Maslennikova, E.
\newblock (2019).
\newblock {ELMo Word Representations For News Protection}.
\newblock In {\em {Working Notes of CLEF 2019 - Conference and Labs of the
  Evaluation Forum}}, number September, pages 9--12, Lugano, Switzerland.

\bibitem[\protect\citename{Moncla \bgroup et al.\egroup
  }2014a]{moncla2014itinerary}
Moncla, L., Gaio, M., and Musti{\`e}re, S.
\newblock (2014a).
\newblock {Automatic Itinerary Reconstruction from Texts}.
\newblock In Matt Duckham, et~al., editors, {\em {Geographic Information
  Science}}, pages 253--267, Cham. Springer International Publishing.

\bibitem[\protect\citename{Moncla \bgroup et al.\egroup
  }2014b]{moncla2014hiking}
Moncla, L., Renteria-Agualimpia, W., Nogueras-Iso, J., and Gaio, M.
\newblock (2014b).
\newblock {Geocoding for Texts with Fine-grain Toponyms: An Experiment on a
  Geoparsed Hiking Descriptions Corpus}.
\newblock In {\em {Proceedings of the 22nd ACM SIGSPATIAL International
  Conference on Advances in Geographic Information Systems}}, SIGSPATIAL '14,
  pages 183--192, New York, NY, USA. ACM.

\bibitem[\protect\citename{{Moreira-Matias} \bgroup et al.\egroup
  }2012]{moreira2012predictive}
{Moreira-Matias}, L., {Gama}, J., {Ferreira}, M., and {Damas}, L.
\newblock (2012).
\newblock {A predictive model for the passenger demand on a taxi network}.
\newblock In {\em {2012 15th International IEEE Conference on Intelligent
  Transportation Systems}}, pages 1014--1019, Sep.

\bibitem[\protect\citename{Normile}2005]{normile426birds}
Normile, D.
\newblock (2005).
\newblock {Are Wild Birds to Blame?}
\newblock {\em Science}, 310(5747):426--428.

\bibitem[\protect\citename{Noulas \bgroup et al.\egroup
  }2011]{noulas2011empirical}
Noulas, A., Scellato, S., Mascolo, C., and Pontil, M.
\newblock (2011).
\newblock An empirical study of geographic user activity patterns in
  foursquare.
\newblock {\em ICWSM}, 11(70-573):2.

\bibitem[\protect\citename{Palm \bgroup et al.\egroup }2015]{Palm2015}
Palm, E.~C., Newman, S.~H., Prosser, D.~J., Xiao, X., Ze, L., Batbayar, N.,
  Balachandran, S., and Takekawa, J.~Y.
\newblock (2015).
\newblock {Mapping migratory flyways in Asia using dynamic Brownian bridge
  movement models}.
\newblock {\em Movement Ecology}, 3(1):3.

\bibitem[\protect\citename{Pedregosa \bgroup et al.\egroup
  }2011]{pedregosa2011scikitlearn}
Pedregosa, F., Varoquaux, G., Gramfort, A., Michel, V., Thirion, B., Grisel,
  O., Blondel, M., Prettenhofer, P., Weiss, R., Dubourg, V., Vanderplas, J.,
  Passos, A., Cournapeau, D., Brucher, M., Perrot, M., and Duchesnay, E.
\newblock (2011).
\newblock {Scikit-learn: Machine Learning in Python}.
\newblock {\em Journal of Machine Learning Research}, 12:2825--2830.

\bibitem[\protect\citename{Pennington \bgroup et al.\egroup
  }2014]{pennington2014glove}
Pennington, J., Socher, R., and Manning, C.~D.
\newblock (2014).
\newblock {GloVe: Global Vectors for Word Representation}.
\newblock In {\em {Empirical Methods in Natural Language Processing (EMNLP)}},
  pages 1532--1543.

\bibitem[\protect\citename{{Perky Pet}}2020]{perkypet2020birds}
{Perky Pet}.
\newblock (2020).
\newblock When do birds migrate?

\bibitem[\protect\citename{Peters \bgroup et al.\egroup }2018]{peters2018deep}
Peters, M., Neumann, M., Iyyer, M., Gardner, M., Clark, C., Lee, K., and
  Zettlemoyer, L.
\newblock (2018).
\newblock {Deep Contextualized Word Representations}.
\newblock In {\em {Proceedings of the 2018 Conference of the North American
  Chapter of the Association for Computational Linguistics: Human Language
  Technologies, Volume 1 (Long Papers)}}, pages 2227--2237.

\bibitem[\protect\citename{Pezanowski and Mitra}2020]{pez2020movementlr}
Pezanowski, S. and Mitra, P.
\newblock (2020).
\newblock Geomovement corpora.

\bibitem[\protect\citename{Pujara \bgroup et al.\egroup
  }2011]{pujara2011reducing}
Pujara, J., London, B., and Getoor, L.
\newblock (2011).
\newblock {Reducing label cost by combining feature labels and crowdsourcing}.
\newblock In {\em {ICML workshop on Combining learning strategies to reduce
  label cost}}, pages 1--5.

\bibitem[\protect\citename{Richter and Klippel}2005]{richter2005context}
Richter, K.-F. and Klippel, A.
\newblock (2005).
\newblock {A Model for Context-Specific Route Directions}.
\newblock In Christian Freksa, et~al., editors, {\em {Spatial Cognition IV.
  Reasoning, Action, Interaction}}, pages 58--78, Berlin, Heidelberg. Springer
  Berlin Heidelberg.

\bibitem[\protect\citename{Rowland \bgroup et al.\egroup
  }1997]{rowland1997starkey}
Rowland, M.~M., Bryant, L.~D., Johnson, B.~K., Noyes, J.~H., Wisdom, M.~J., and
  Thomas, J.~W.
\newblock (1997).
\newblock {The Starkey project: history, facilities, and data collection
  methods for ungulate research}.
\newblock {\em {Gen. Tech. Rep. PNW-GTR-396. Portland, OR: US Department of
  Agriculture, Forest Service, Pacific Northwest Research Station. 62 p}}, 396.

\bibitem[\protect\citename{Schneider}2016]{ibm2016unstructured}
Schneider, C.
\newblock (2016).
\newblock The biggest data challenges that you might not even know you have,
  May.

\bibitem[\protect\citename{{Soares Junior} \bgroup et al.\egroup
  }2017]{Soares2017}
{Soares Junior}, A., Renso, C., and Matwin, S.
\newblock (2017).
\newblock {ANALYTiC: An Active Learning System for Trajectory Classification.}
\newblock {\em IEEE computer graphics and applications}, 37(5):28--39.

\bibitem[\protect\citename{Stock \bgroup et al.\egroup }2013]{stock2013corpus}
Stock, K., Pasley, R.~C., Gardner, Z., Brindley, P., Morley, J., and Cialone,
  C.
\newblock (2013).
\newblock {Creating a Corpus of Geospatial Natural Language}.
\newblock In Thora Tenbrink, et~al., editors, {\em {Spatial Information
  Theory}}, pages 279--298, Cham. Springer International Publishing.

\bibitem[\protect\citename{Wallgr{\"u}n \bgroup et al.\egroup
  }2018]{wallgrun2018geocorpora}
Wallgr{\"u}n, J.~O., Karimzadeh, M., MacEachren, A.~M., and Pezanowski, S.
\newblock (2018).
\newblock {GeoCorpora: building a corpus to test and train microblog
  geoparsers}.
\newblock {\em {International Journal of Geographical Information Science}},
  32(1):1--29.

\bibitem[\protect\citename{Wikelski and Kays}2019]{wikelski2019}
Wikelski, M. and Kays, R.
\newblock (2019).
\newblock Movebank: archive, analysis and sharing of animal movement data.
\newblock Hosted by the Max Planck Institute for Ornithology.

\bibitem[\protect\citename{{Zheng} \bgroup et al.\egroup
  }2008]{zheng2008geolife}
{Zheng}, Y., {Wang}, L., {Zhang}, R., {Xie}, X., and {Ma}, W.
\newblock (2008).
\newblock {GeoLife: Managing and Understanding Your Past Life over Maps}.
\newblock In {\em {The Ninth International Conference on Mobile Data Management
  (mdm 2008)}}, pages 211--212, Apr.

\end{thebibliography}

\end{document}